%% file: marks2024iros.tex
\documentclass[letterpaper, 10 pt, conference]{ieeeconf}
\IEEEoverridecommandlockouts    %
\input{stachnisslab-latex}

\usepackage{amsmath}
\usepackage{hyperref}
\usepackage{makecell}
\usepackage{multirow}

\newcommand{\DatasetName}{BonnBeetClouds3D\xspace}

\title{\LARGE \bf \DatasetName: A Dataset Towards Point Cloud-Based Organ-Level Phenotyping of Sugar Beet Plants Under Real Field Conditions}

\renewcommand{\and}{\hspace{0.7cm}}

\author{Elias Marks \and Jonas Bömer \and Federico Magistri \and  Anurag Sag \and Jens Behley \and Cyrill Stachniss%
  \thanks{E. Marks, F. Magistri, A. Sag, J. Behley, and C. Stachniss are with the Center for Robotics, University of Bonn, Germany. 
  J. Bömer is with Institute of Sugar Beet Research, Göttingen, Germany.
  Stachniss is additionally with the Lamarr Institute for Machine Learning and Artificial Intelligence, Germany.}%
  \thanks{This work has partially been funded 
  by the Deutsche Forschungsgemeinschaft (DFG, German Research Foundation) under Germany's Excellence Strategy, EXC-2070 -- 390732324 -- PhenoRob,
  under STA~1051/5-1 within the FOR 5351~(AID4Crops), and
  by the Federal Ministry of Food and Agriculture~(BMEL) based on a decision of the Parliament of the Federal Republic of Germany via the Federal Office for Agriculture and Food~(BLE) under the innovation support programme under funding no~28DK108B20~(RegisTer).
  }%
}

\begin{document}
\maketitle
\thispagestyle{empty}
\pagestyle{empty}

\begin{abstract}
  Agricultural production is facing challenges in the next decades induced by climate change and the need for more sustainability by reducing its impact on the environment. Advances in field management through robotic intervention, monitoring of crops by autonomous unmanned aerial vehicles (UAVs) supporting breeding of novel and more resilient crop varieties can help to address these challenges. The analysis of plant traits is called phenotyping and is an essential activity in plant breeding; it however involves a great amount of manual labor. With this paper, we provide means to better tackle the problems of instance segmentation to support robotic intervention and automatic fine-grained, organ-level geometric analysis needed for precision phenotyping. As the availability of real-world data in this domain is relatively scarce, we provide a novel dataset that was acquired using UAVs capturing high-resolution images of real breeding trials containing 48 plant varieties and therefore covering a relevant morphological and appearance spectrum. This enables the development of approaches for instance segmentation and autonomous phenotyping that generalize well to different plant varieties. Based on overlapping high-resolution images taken from multiple viewing angles, we provide photogrammetric dense point clouds and provide detailed and accurate point-wise labels for plants, leaves, and salient points as the tip and the base in 3D. Additionally, we include measurements of phenotypic traits performed by experts from the German Federal Plant Variety Office on the real plants, allowing the evaluation of new approaches not only on segmentation and keypoint detection but also directly on actual traits. The provided labeled point clouds enable fine-grained plant analysis and support further progress in the development of automatic phenotyping approaches, but also enable further research in surface reconstruction, point cloud completion, and semantic interpretation of point clouds.
\end{abstract}

\section{Introduction}
\label{sec:intro}

\begin{figure}[t]
  \centering
  \includegraphics[width=1.0\linewidth]{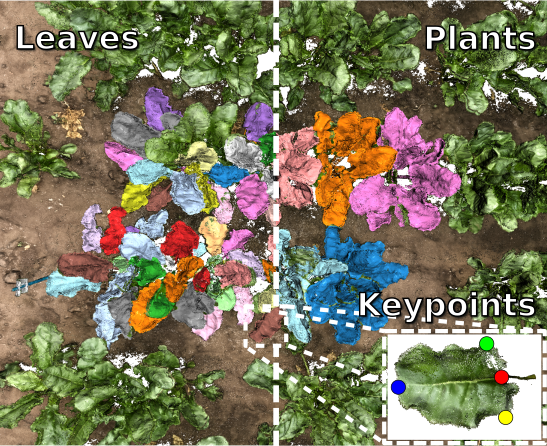}
  \caption{Our dataset, called \DatasetName consists of high-resolution point clouds extracted via bundle adjustment of high-resolution images captured from multiple viewing angles by a UAV. 
  Together with the point clouds, we also provide accurate per-point labels of the plants, leaves, and salient points of the leaves, such as leaf tips, corners and bases. Furthermore, we provide reference values of measurements that are relevant for phenotyping, such as leaf lenghts and widths.}
  \label{fig:motivation}
\end{figure}

Agricultural production of food, feed, and fiber is facing big challenges in the next decades.
While the world population is steadily growing, inducing a higher demand for agricultural products, the availability of arable land is decreasing due to climate change~\cite{pravalie2021er}.
At the same time, we have to reduce the effects of conventional agricultural practices on our environment calling for a more sustainable agricultural production in order to preserve biodiversity~\cite{steinbachinger2021oa} and the productivity of arable lands~\cite{wortman2012rafs, maeder2002science, diacono2010asd}.
As one measure, agricultural production needs to significantly reduce the amount of applied agrochemicals, such as pesticides and herbicides.

A potential pathway to enable a more sustainable agricultural production is the breeding of more resilient crop varieties, novel field management techniques including more targeted interventions at a plant level, and an increase of non-chemical weeding practices to reduce the amount of needed herbicides.
These areas benefit from artificial intelligence-based solutions, especially advances in robotics and computer vision~\cite{sparrow2021robots} to obtain a better estimate of the crop status in the field.
In particular, the development of more capable perception systems for robotic platforms or automatized tractors could lead to a higher automation enabling plant-specific treatments, but also novel ways of analyzing data from the field could lead to advancement towards more sustainable agricultural production.

In this paper, we provide a dataset to target plant trait assessment in the field by observing the phenotype of a plant, which is an expression of its genotype, influenced by environmental conditions, and affected by the field management. An example of the provided data can be seen in~\figref{fig:motivation}.

\begin{figure}[t]
  \centering
  \includegraphics[width=1\linewidth]{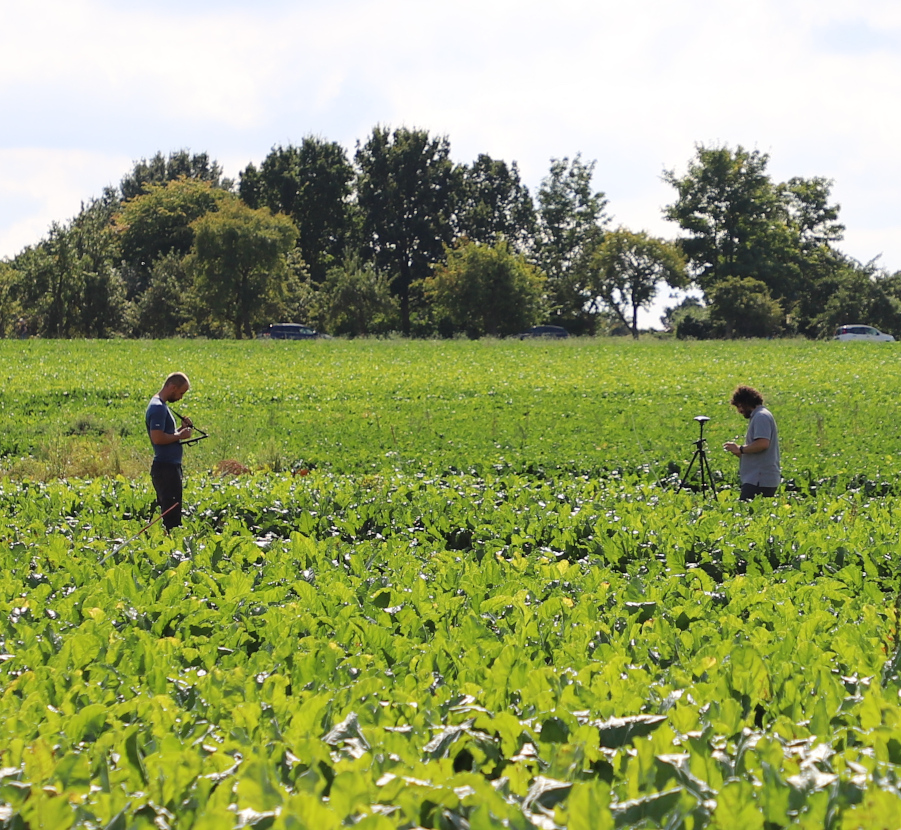}
  \caption{Experts estimating phenotypic traits in the fields. This task is performed by visual inspection and, thus, depends also on the subjective observation of the experts.}
  \label{fig:expert_phenotyping}
\end{figure}

The assessment of the morphological characteristics of a plant is to this date often performed by manual measurements or visual scoring of a trained expert. These experts evaluate different traits such as growth stage, leaf attitude, plant height, and leaf length as shown in~\figref{fig:expert_phenotyping}.
However, this assessment is a tedious and labour-intensive task, and also subjective to a certain extent. All this affects the spatial extent, reliability, and repeatability of phenotyping activities.

A potential solution to the aforementioned limitations of conventional phenotyping is the use of algorithmic solutions to automate the process, which also leads to repeatable and objective measurements of phenotypic traits.
In particular, a combination of a robotic solution, which automatically acquires the relevant data in combination with a robot perception or computer vision-based approach to extract such traits from the recorded sensor data is a way to tackle the so-called "phenotyping bottleneck"~\cite{furbank2011ps}.

The significant progress in image-based perception, but also the development of increasingly capable perception systems in autonomous driving, has been propelled by the availability of domain-specific datasets~\cite{geiger2012cvpr,lin2014eccv,cordts2016cvpr,behley2019iccv,sun2020cvpr,chang2019cvpr-ataf} to a large extent.
While there has been recently an increasing interest in agricultural robotics, this progress is stifled by the availability of data to study and advance perception in this domain.
There is an increasing number of datasets available in the agricultural domain for semantic interpretation of 2D field images~\cite{chebrolu2017ijrr, sa2018rs, weyler2024tpami}, not many, however, contain 3D plant data, which is highly relevant for morphological trait estimation~\cite{schunck2021plosone}.

The main contribution of this paper is a large dataset covering 48 different varieties of sugar beets that we recorded on real breeding trials with over 3,000 plants, with point-wise annotations for 186 individual plants and 2,661 individual leaves enabling the development and evaluation of segmentation, detection and actual in-field phenotyping algorithms.
To allow for direct evaluation of tasks such as extraction of phenotypic traits, we also provide reference values for commonly evaluated traits such as leaf length, leaf width and stem length alongside the position of over 10,000 salient points such as the leaf tips, leaf corners and plant centers.
To obtain the data, data loading utilities, evaluators, or submit your results to the public challenge please visit \href{https://bonnbeetclouds3d.ipb.uni-bonn.de/}{https://bonnbeetclouds3d.ipb.uni-bonn.de}

In sum we make the following contributions.
Our dataset enables:
(i) studying phenotyping using photogrammetric point clouds by exploiting geometric information;
(ii) instance segmentation of plant organs using 3D point clouds;
(iii) estimation of key locations on the leaves that are useful for phenotyping.

\section{Related Work}
\label{sec:related}

There has been increasing interest in semantic interpretation of field images targeting semantic~\cite{lottes2016icra,lottes2017icra,lottes2017jfr,lottes2020jfr,milioto2018icra}, instance~\cite{weyler2022wacv}, and panoptic segmentation~\cite{roggiolani2022icra} of crops and weeds.
These approaches mainly use large image datasets with pixel-wise annotations of crops and weeds~\cite{chebrolu2017ijrr,sa2018rs}, but also leaf annotations~\cite{kierdorf2022jfr, weyler2024tpami}. 
For a broader overview of agricultural datasets for semantic interpretation, we refer to the survey by Lu \etal~\cite{lu2020cea}.

While there are several image datasets available, few datasets provide point cloud data revealing the geometric structure of the plants that seems mandatory for a fine-grained structural analysis of phenotypic traits. 
Chaudhury~\etal~\cite{chaudhury2020eccvws} proposed a synthetic dataset generated based on plant models with the addition of noise.
While being able to generate large amounts of data, the gap between simulation and reality poses a challenge for the deployment of approaches developed on this data to real fields.
The Pheno4D dataset~\cite{schunck2021plosone} provides point clouds of real tomato and maize plants at different growth stages captured with a high-precision laser scanner with accurate labels of individual plant organs.
In contrast to our dataset, these point clouds were acquired in a lab setting under controlled conditions.
Dutagaci~\etal~\cite{dutagaci2020pm} instead provide a dataset with real world data, but it contains rose plants.
Khanna~\etal~\cite{khanna2019pm} provide a dataset containing sugar beet plants, but they do not provide point-wise instance labels nor reference measurements for leaf measures.
\tabref{tab:dataset_compar} summarizes the key characteristics of the point cloud datasets in the agricultural domain, which shows that our dataset has unique characteristics not covered by the available datasets.
Our dataset covers real breeding trials showing real-world lighting conditions, but also occlusions induced by the planting of the crops.
Furthermore, it is augmented with real measurements by the phenotyping experts from the German Federal Plant Variety Office providing reference measurements of per-plot phenotypic traits.

\begin{table*}
  \centering
  \resizebox{.9\linewidth}{!}{
  \begin{tabular}{ccccccc}
  \toprule
  \textbf{Dataset} & \textbf{Real data} & \textbf{Field} & \textbf{Semantic labels} & \textbf{Instance labels} & \textbf{Phenotypic measurements} & \textbf{Varieties}\\
  \midrule
  Dutagaci~\etalcite{dutagaci2020pm} & \checkmark &  & \checkmark & \checkmark &  & Rose \\
  Khanna~\etalcite{khanna2019pm} & \checkmark & \checkmark &  &  &  & Sugar beet \\
  Pheno4D~\cite{schunck2021plosone}      & \checkmark &  & \checkmark & \checkmark &  & Tomato, Maize \\
  \textbf{BonnBeetClouds3D (Ours)} & \checkmark & \checkmark & \checkmark & \checkmark & \checkmark & Sugar beet \\
  \bottomrule
  \end{tabular}
  }
  \caption{Overview of the key characteristics of the point cloud datasets in the agricultural domain, which shows that our dataset has unique characteristics not covered by the available datasets in the agricultural domain.}
  \label{tab:dataset_compar}
\end{table*}

\section{The \DatasetName Dataset}
\label{sec:main}

\begin{figure}[t]
  \centering
  \includegraphics[width=0.95\linewidth]{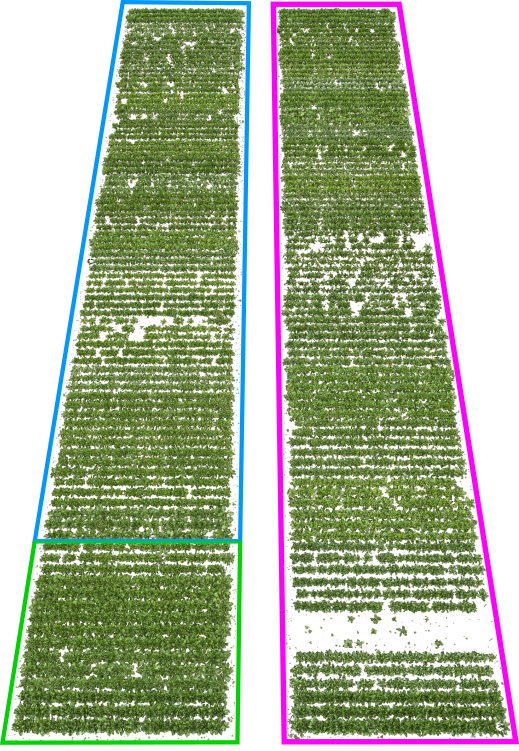}
  \caption{Overview of the captured field. The data used for training of deep learning approaches is highlighted in purple, the validation area is highlighted in green and the test data is in the blue rectangle.}
  \label{fig:captured_data}
\end{figure}

\subsection{Field Setup}
We recorded the data of the sugar beet trials of the Federal Plant Variety Office in Magdeburg, Germany (N52.112028°, E011.558448°).
The trial was set up in a single factorial block design with the factor variety which contained two repetitions.
The plot size was 1.5\,m by 7.0\,m and contained three rows of plants with a row distance of $0.5$\,m.
We cover a total of 48 varieties in our dataset.

The field technicians of the German Federal Plant Variety Office sowed the plants at a distance of $6$\,cm inside the row and later manually thinned them out to a final distance of 18\,cm by manual hoeing.
Plant protection products like herbicides were used in the normal course of business. Any remaining weeds were regularly removed by manual hoeing.
In addition, the experiment was artificially irrigated when the weather conditions were too dry. These labor-intensive tasks were made to create ideal conditions for the development of the varieties.

\subsection{Sensor Setup}

We recorded our field data using a UAV equipped with a PhaseOne iXM-100 camera with a $80\,$mm RSM prime lens mounted on a gimbal to obtain {motion-stabilized} RGB 100\,MP images at a resolution of $11,664\,\text{px}\times 8,750\,\text{px}$ per image. 
The UAV was flying at a height of approx.~$21\,$m, resulting in a ground sampling distance~(GSD) of $1$\,$\frac{\text{mm}}{\text{px}}$.
We flew three missions on the same field with a camera angle of $45^{\circ}$, $90^{\circ}$, and $135^{\circ}$ degrees from the ground plane, which allows us to obtain good coverage of the crops including the lower parts even in the tight row spacing.

\subsection{Data Processing and Labeling}

We process the recorded, overlapping 100 megapixel raw camera images of different view angles to determine the relative poses between the images via photogrammetric bundle adjustment, where we used the GNSS positions as initial camera positions.
Using the poses, we compute a 3D point cloud of the complete field as shown in \figref{fig:captured_data}.

\begin{figure}[t]
  \centering
  \includegraphics[width=0.95\linewidth]{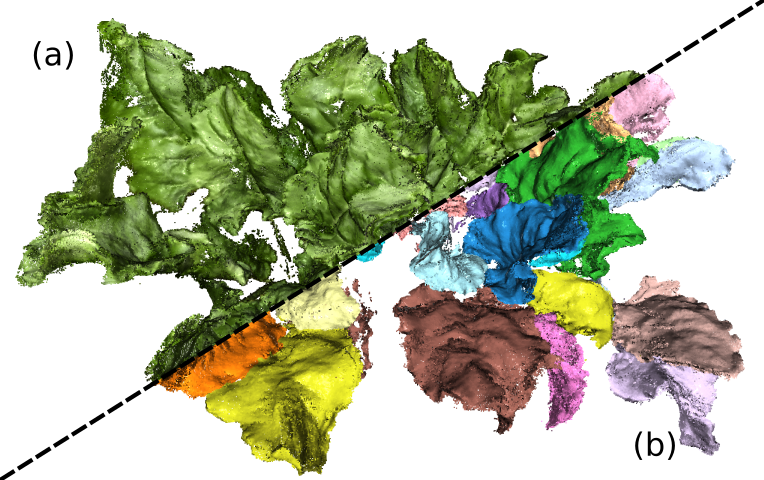}
  \caption{In the top left part (a) we show an unlabeled point cloud with the original leaf colors, while in the bottom right (b) we show the color-coded labeled leaf instances, where different colors correspond to different leaves.}
  \label{fig:label_sample}
\end{figure}

To obtain a reliably accurate segmentation of plants and leaves, we manually labeled the individual plants and leaf instances in the point clouds using the Semantic Segmentation Editor by Hitachi Automotive and Industrial Laboratory.
An example of the annotated point cloud is shown in~\figref{fig:label_sample}. 

\begin{figure*}[t]
  \centering
  \includegraphics[width=0.99\linewidth]{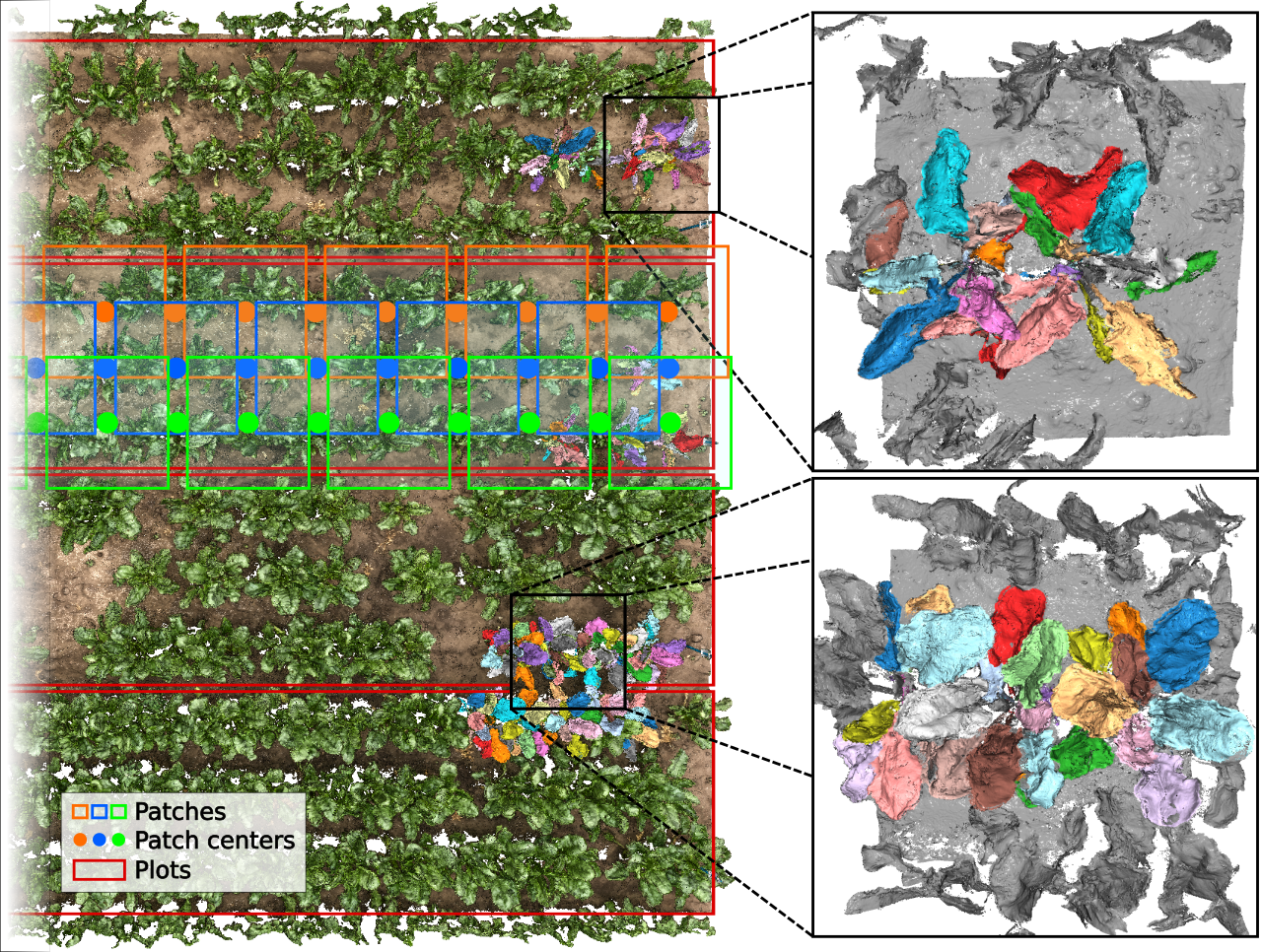}
\caption{The patch extraction process from the point cloud of the field. As can be seen in the field representation, a subset of plant has been annotated, here the leaf color encode the respective leaf instances. From the whole field point cloud we then extract patches, as the entire point cloud is too big for being processed at once. The colored dots represent the centers of all the patches extracted from that plot. The colored boxes, instead, show only a subset of the extracted patches for better visualization. On the right there are two example patches with color-coded leaf ids and grey for the unlabeled part.}
\label{fig:patching}
\end{figure*}
\begin{figure}[t]
  \centering
  \includegraphics[width=0.9\linewidth]{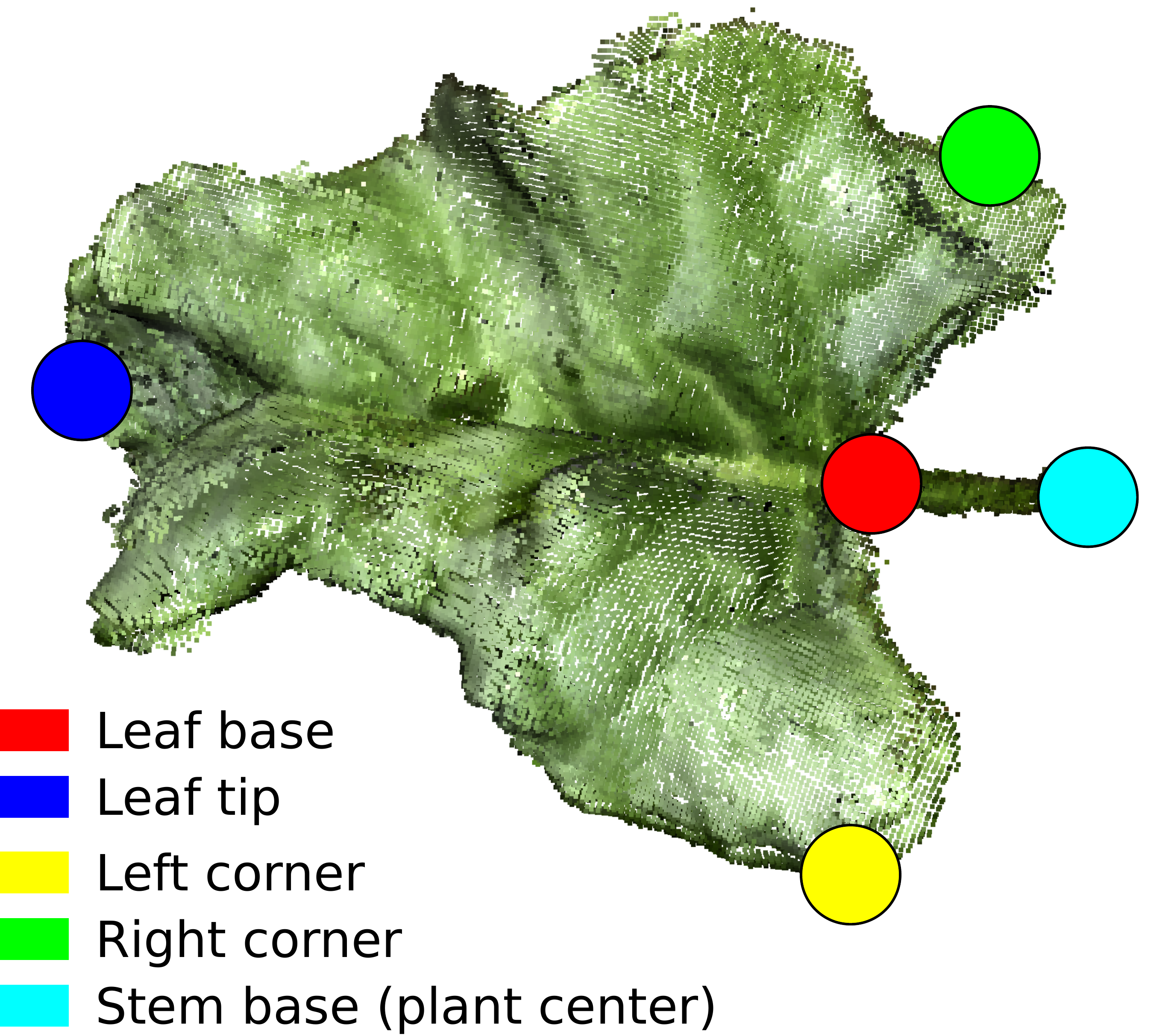}
  \caption{Labeled keypoints. The leaf base is the point where the stem joins the leaf blade, the left and right corners are the two points furthest apart from the main leaf axis, which therefore can be used for the computation of the maximum leaf width, the leaf tip is the topmost point of the leaf and the stem base is the point where the stem joins the beet crown, which we also define as the plant center.}
  \label{fig:keypoint_fig}
\end{figure}

The annotated plants were carefully checked by a person different from the one that annotated them in the first place, so we ensured that multiple people checked each label in order to obtain a higher accuracy in the resulting annotations.
There are many plants present for every variety, making it intractable and also unnecessary to label all of them. 
Therefore, we selected one group of adjacent plants per variety and annotated them. 
As processing the whole field point cloud at once is impossible due to the amount of points, we extracted regularly spaced and overlapping patches covering the crop rows.
The patching process is shown in~\figref{fig:patching}.
The whole labeling process required substantial manual labor, but resulted in 186 annotated plants and 2,661 annotated leaves, covering all 48 varieties in the breeding trial.
We furthermore manually annotated different relevant keypoints of each individual leaf as shown in \figref{fig:keypoint_fig}. The leaf base represents the point where the leaf blade joins the petiole, the leaf tip represent the point that is furthest from the petiole and the lateral corners represent the points of the leaf that are furthest away from the central axis of the leaf. The stem base, instead, is the point where the stem joins the beet crown, which we also define as the plant center.
In sum, we annotated over 10,000 such keypoints.

The German Federal Plant Variety Office performed measurements of different variety traits of 30 leaves for every breeding plot.
These are the leaf blade length and width, and the petiole length and width. 
Additionally, they provided us the average leaf angle with respect to the ground, average plant height, the curliness and the curviness of the leaf border for each plot, which was scored by trained phenotyping experts.

\begin{table*}[t]
  \caption{Attributes of the points in the dataset. Each point has an associated coordinate, normal, color, leaf id, plant id, plot id and keypoint id.}
  \label{tab:attributes}
  \centering
  \begin{tabular}{cccc} 
    \toprule
    \textbf{Attributes} & \textbf{Data Type} & \textbf{Range} & \textbf{Notes}\\
    \midrule
    point coordinates & $\mathbb{R}^{\text{N}\times3}$ & [$-\infty$ ,$\infty$] & Point coordinates in m\\ 
    point normal & $\mathbb{R}^{\text{N}\times3}$ & [$-1$ ,$1$] & Unit vector representing point normal\\ 
    point color & $\mathbb{Z}^{\text{N}\times3}$ & [0,255] & 8 bit point color\\ 
    leaf id & $\mathbb{Z}^{\text{N}\times1}$ & [-2, number of leaves] & -2: unlabeled, -1: plant center\\ 
    plant id & $\mathbb{Z}^{\text{N}\times1}$ & [-1, number of plants] & -1: unlabeled\\ 
    plot id & $\mathbb{Z}^{\text{N}\times1}$ & [0, number of plots] & Number of the plot the point belongs to\\ 
    keypoint id & $\mathbb{Z}^{\text{N}\times1}$ & [-1,3] & -1: no keypoint, 0: base, 1: left corner, 2: tip, 3: right corner\\ 
    \bottomrule
  
  \end{tabular}
 \end{table*}
\subsection{Data Organization}

To ensure a fair evaluation of the tested approaches, we divided the dataset into a training set consisting of 1,782 leaves from 128 plants to develop and train approaches, a validation set of 260 leaves from 17 plants to tune the hyperparameter of the approaches and a test set of 619 leaves from 41 plants to validate the performance of the developed approaches.
We provide labels for the train and validation sets. 
The point clouds are divided into patches of 1\,m by 1\,m by 1\,m to make them easy to use for training deep neural networks. We use the ply format to provide these patches, where each point has the attributes shown in \tabref{tab:attributes}.

The keypoint ids are labeled as subsets of the leaves, with multiple points of the same leaf having the same keypoint id. To get the most precise keypoint position, we find the centroid all points of one leaf having the same keypoint id.
The ground truth measurements of the the phenotypic traits are delivered for every plot in csv-format. These include:
\begin{itemize}
  \item leaf blade length
  \item leaf blade width
  \item petiole length
  \item petiole width
\end{itemize}

\begin{figure*}[t]
  \centering
  \includegraphics[width=0.8\linewidth]{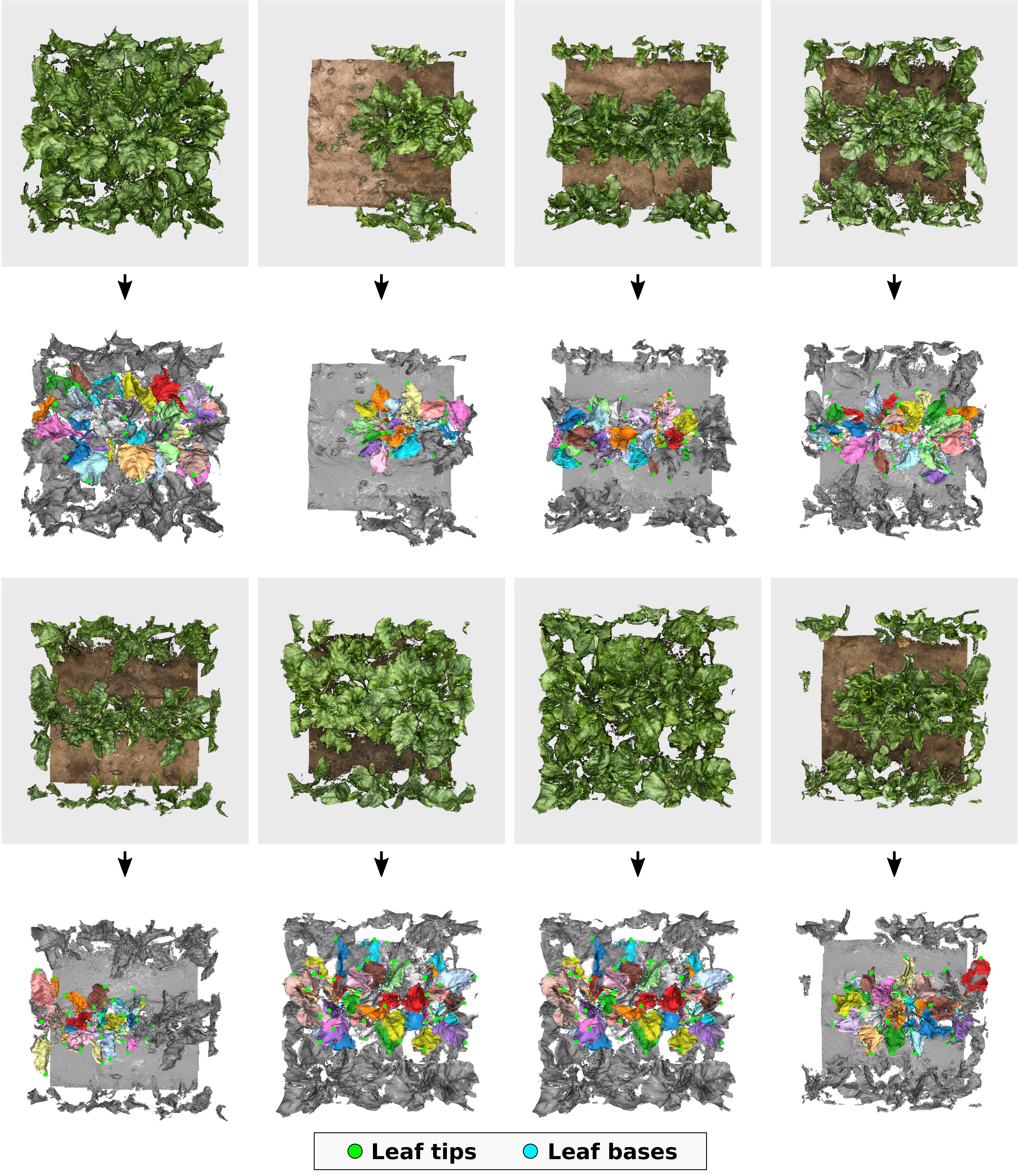}
\caption{The point cloud patches, shown both with RGB colors and color-coded leaf ids based on the annotations. The grey points represent the unlabeled part. The colored circles in the annotated view represent the position of the annotated keypoints. To avoid the figure to be too cluttered we show only the leaf tips and bases and omit the two corners.}
\label{fig:data_samples}
\end{figure*}

We show some examples of the data in~\figref{fig:data_samples}.

To obtain the data, data loading utilities, evaluators, or submit your results to the public challenge please visit \href{https://bonnbeetclouds3d.ipb.uni-bonn.de}{https://bonnbeetclouds3d.ipb.uni-bonn.de}

\section{Experiments}
In sum our dataset supports:
(i) studying phenotyping using photogrammetric 3D point clouds, by exploiting geometric information;
(ii) instance segmentation of plant organs using 3D point clouds;
(iii) estimation of key locations on the leaves that are useful for phenotyping.

We show how different existing approaches perform on the different tasks above, so that they can be used as a baseline in the evaluation of new approaches.

 \begin{table}
  \caption{Leaf trait detection in the field. In the table we report the errors in the estimation of the 3 leaf traits in field conditions. We present the offsets in millimeters of the estimations from the manually measured ground truth values for each approach.}
  \label{tab:pheno_exp}
  \centering
 \setlength\tabcolsep{0pt}
\begin{tabular*}{0.95\linewidth}{@{\extracolsep{\fill}} cccc }
  \toprule
  \multirow{3}{*}{\textbf{Approach}} & \multirow{2}{*}{\textbf{\makecell{leaf\\length}}}\multirow{2}{*}{\hspace{1mm}\makecell{[mm]}} & \multirow{2}{*}{\textbf{\makecell{blade\\length}}}\multirow{2}{*}{\hspace{1mm}\makecell{[mm]}} & \multirow{2}{*}{\textbf{\makecell{blade\\width}}}\multirow{2}{*}{\hspace{1mm}\makecell{[mm]}} \\
  & & & \\
  & $\downarrow$ avg  & $\downarrow$ avg  & $\downarrow$ avg \\
 \midrule
 CPD~\cite{myronenko2010pami}                       & 205                 & 139                    & 109       \\
 PF--SGD~\cite{marks2022icra}                   & \textbf{35}                 & 52                    & 15        \\
 CurvLeaf~\cite{marks2024cea}                      & 38            & \textbf{18}               & \textbf{13}    \\
 \bottomrule

\end{tabular*}
\end{table}

\begin{table}
  \caption{Instance segmentation of plant organs. In the table we report the performance of the three baselines. We report panoptic quality (PQ), segmentation quality (SQ), and recognition quality (RQ). The metrics are introduced by Kirillov~\etalcite{kirillov2019cvpr-ps}.}
  \label{tab:segmentation_exp}
  \centering
  \setlength\tabcolsep{0pt}
  \begin{tabular*}{0.95\linewidth}{@{\extracolsep{\fill}} cccc }
  \toprule
  \multirow{2}{*}{\textbf{Approach}} & \textbf{PQ} [\%] & \textbf{SQ} [\%] & \textbf{RQ} [\%] \\
   & $\uparrow$ & $\uparrow$ & $\uparrow$\\
  \midrule
  SoftGroup~\cite{thang2022cvpr}   & 72.07 & 79.45 & 90.69 \\
  Mask3D~\cite{schult2023icra}     & 18.47 & 67.71 & 25.80 \\
  HP-LIS~\cite{marks2023ral}      & \textbf{75.58} & \textbf{80.97} & \textbf{93.17} \\
  \bottomrule

\end{tabular*}
\end{table}

\subsection{Estimating Geometric Phenotypic Traits in the Fields}
As we provide precise ground truth measurements for the geometric phenotypic traits, i.e., leaf length, leaf blade length, and leaf width, the dataset can be used to evaluate how well new approaches are able to extract those features from the photogrammetric point clouds.
This kind of approaches are important for improving the throughput and accuracy of phenotyping practices.
We evaluated the approaches based on the prediction error for the different traits in mm.
The results of the three approaches are shown in~\tabref{tab:pheno_exp}.

The first approach consists in using the coherent point drift algorithm (CPD)~\cite{myronenko2010pami} to deform a triangular mesh to the point cloud and estimating the leaf parameters based on the obtained deformed mesh.
The algorithm reconstructs completely observed leaves quite well, but it fails for the ones presenting occlusions. The mesh resulting from CPD always collapses on the visible part of the leaves. This leads to a bad performance in the estimation of the leaf dimensions.

Another approach that suffers less from this issue is the partiality filtered stochastic gradient descent, denoted as PF-SGD~\cite{marks2022icra}. Here we used gradient descent to optimize the position of the vertices of a triangular mesh that represents our leaf model. The vertex positions are optimized in order to deform the mesh onto the point cloud of the leaf, and then the leaf dimensions are measured on the deformed model.
This approach improves substantially in comparison to CPD.

To improve the robustness of the reconstruction we developed a curve based leaf model fitting algorithm, denoted as CurvLeaf~\cite{marks2024cea}. Instead of optimizing the position of individual vertices, we here optimize the parameters of parametric curves to best fit the leaf model to the point clouds. 
This has the advantage that the deformations can be better constrained, leading to more robust estimation of leaf lengths, widths and blade lengths.

 \begin{table*}[t]
  \caption{Detection of leaf keypoints. In the table we report the errors in the estimation of the 4 leaf keypoints in field conditions. We present the offsets in millimeters of the estimations from the manually annotated ground truth keypoints.}
  \label{tab:kp_exp}
  
  \centering
  \resizebox{.7\linewidth}{!}{
  \begin{tabular}{c c c c c} 
    \toprule
    \multirow{2}{*}{\textbf{Approach}} & \textbf{leaf tip} [mm] & \textbf{leaf base} [mm] & \textbf{left corner} [mm] & \textbf{right corner} [mm]\\
     & $\downarrow$ avg  & $\downarrow$ avg  & $\downarrow$ avg & $\downarrow$ avg\\
    \midrule
    SVD                                   & \textbf{15}                 & 33                    & \textbf{43}        & 46\\
    CPD~\cite{myronenko2010pami}          & 30                 & 68                    & 52        & 50\\
    PF--SGD~\cite{marks2022icra}                   & 17                 & 24                    & \textbf{43}         & \textbf{41}\\
    CurvLeaf~\cite{marks2024cea}                      & 21            & \textbf{19}              & 49     & 48 \\
    \bottomrule
  
  \end{tabular}
  }
 \end{table*}

\subsection{Instance Segmentation of Plant Organs using 3D Point Clouds}
Another important task needed for autonomous high throughput phenotyping is the detection and segmentation of individual plant organs such as leaves and stems in the field. Only after detecting which part of the the point cloud corresponds to a specific leaf one can estimate the leaf dimensions.
We compared different approaches for instance segmentation in point clouds that can be used as baselines for the evaluation of new approaches.
As metric for comparison we used panoptic quality~\cite{kirillov2019cvpr-ps}, which is defined as:
\begin{equation}
  \label{eq:PQ}
  \text{PQ} = \underbrace{\frac{\sum_{(p,g) \in \text{TP}} \text{IoU}(p,g)}{|\text{TP}|}}_{\text{SQ}} \underbrace{\frac{|\text{TP}|}{|\text{TP}| + \frac{1}{2}|\text{FP}| + \frac{1}{2}|\text{FN}|}}_{\text{RQ}}.
  \end{equation}
The performance in terms of panoptic quality~\cite{kirillov2019cvpr-ps} is shown in~\tabref{tab:segmentation_exp}.

The first approach that we evaluated is SoftGroup~\cite{thang2022cvpr}, which is a geometric learning approach that predicts offset vectors pointing to the corresponding leaf center for each point of the point cloud. These offset vectors are then clustered to obtain sets of point corresponding to individual leaves.
The resulting leaves are separated quite well, but there are some points assigned to the wrong leaf, especially in the border regions.

To cover different kinds of neural networks, we evaluated a query-based transformer approach, which is performing very well on existing scene segmentation datasets like ScanNet~\cite{dai2017cvpr}, S3DIS~\cite{armeni2016cvpr}, and SemanticKITTI~\cite{behley2019iccv}.
The approach by Schult~\etalcite{schult2023icra}, called Mask3D, learns query vectors during the training phase that are then decoded by the transformer decoder heads into masks covering single instances.
Mask3D's performance suffers a lot from the fact that labels are not present for all points in the point cloud patches while training, leading to worse performance then the other baselines.

Finally, we evaluated a domain-specific approach~\cite{marks2023ral}, denoted as HP-LIS. As SoftGroup, it estimates offset vectors pointing to the centroid of the corresponding leaf for each point and cluster them into leaf instances. We however use KPConv~\cite{thomas2019iccv} as backbone, which implements a convolution defined directly in continuous space, instead of voxelizing the point cloud into a 3D grid. For clustering we use HDBSCAN~\cite{mcinnes2017icdmw} which is slower than the clustering algorithm used in SoftGroup but leads to better segmentation performance.

\subsection{Estimation of Key Locations on the Leaves}
The identification of key locations on the leaves such as the leaf tip, the leaf base and the leaf lateral extremities are useful to detect parts of plant organs. These key locations are also useful for computing phenotypic traits relying on them, e.g., the distance between different parts or the relative orientation in space.
This problem can be approached in different ways. One such way is direct detection of the salient points, which we evaluated by Eigenvector analysis of the point cloud, which we report as SVD.
Additionally, we detected the keypoints based on the models fitted by CPD, PF-SGD and CurvLeaf, as those points are known positions on the predefined model.

To compare the performance of the different leaf keypoint detection approaches we compute the L2 distance in mm between the predicted and the manually labeled keypoints, which we show in~\tabref{tab:kp_exp}.
The leaf tips and bases are detected well by some approaches, while the corners are harder to detect. The main reasons for this are the heavy occlusions and irregular leaf shapes. CPD performs the worst for all keypoints, as the meshes are not well fitted to the point clouds. SVD works quite well, especially for the leaf tip detection. CurvLeaf is best at detecting the leaf base, while PF-SGD best detects the corners.

\section{Conclusion}
\label{sec:conclusion}

This paper presents a novel dataset for agricultural robotics, which enables research on automated morphological parameter extraction for phenotyping. 
We provide reference values both for intermediate tasks like instance segmentation and for downstream tasks like performing automated leaf measurements.
Additionally, we evaluated the performance of existing methods, which can be used to solve the different tasks, to provide an insight on the current state of the art and to provide baselines for further research on the topic.
A vision system that is able to segment individual plant organs can enable highly automated robotic interaction with the plants in real fields, but also improve grasping performance, view planning, and mapping.

\section*{Acknowledgments}
We thank the German Plant Variety Office (Bundessortenamt), especially Dr. Manthey, Dr. Lichthardt, Dr. Beukert, Dr. Borg, Ms. Zickenrott, and Mr. Bauske, for maintaining the experimental fields on which the data was collected, for supporting the data collection, for collecting the reference data for the phenotypic traits, for the technical advice on the current phenotyping practices, and for a pleasant collaboration.
We also thank Facundo Ispizua Yamati and Dirk Koops for providing a picture of them while estimating phenotypic traits in the field.

\bibliographystyle{plain_abbrv}

\bibliography{glorified,new}

\end{document}

%% file: stachnisslab-latex.tex
\usepackage{graphics}           
\usepackage{times}              
\usepackage{amsmath}            
\usepackage{amssymb}            
\usepackage{graphicx}
\usepackage{algorithm}
\usepackage[noend]{algpseudocode}
\usepackage{booktabs}
\usepackage{color}
\usepackage[nocompress]{cite} %
\definecolor{instructioncolor}{rgb}{.5,.5,.5}

\usepackage[font=small]{caption}

\def\figref#1{Fig.~\ref{#1}}
\def\tabref#1{Tab.~\ref{#1}}
\def\eqref#1{Eq.~(\ref{#1})}

\makeatletter
\usepackage{xspace}
\DeclareRobustCommand\onedot{\futurelet\@let@token\@onedot}
\def\@onedot{\ifx\@let@token.\else.\null\fi\xspace}

\def\etal{{et al}\onedot}
\makeatother

\def\etalcite#1{\etal~\cite{#1}}

\usepackage{array}
\newcolumntype{L}[1]{>{\raggedright\let\newline\\\arraybackslash\hspace{0pt}}m{#1}}
\newcolumntype{C}[1]{>{\centering\let\newline\\\arraybackslash\hspace{0pt}}m{#1}}
\newcolumntype{R}[1]{>{\raggedleft\let\newline\\\arraybackslash\hspace{0pt}}m{#1}}